\documentclass[runningheads]{llncs}

\usepackage{eccv}

\usepackage{eccvabbrv}

\usepackage{graphicx}
\usepackage{booktabs}

\usepackage[accsupp]{axessibility}  

\usepackage{algorithm}
\usepackage{algpseudocode}

\usepackage{hyperref}

\usepackage{orcidlink}

\begin{document}

\title{Tensorial template matching for fast cross-correlation with rotations and its application for tomography} 


\author{Antonio Martinez-Sanchez\inst{1,*}\orcidlink{0000-0002-5865-2138} \and
Ulrike Homberg\inst{2}\orcidlink{0009-0003-0045-0861} \and
José María Almira\inst{3}\orcidlink{0000-0001-7200-4076} \and
Harold Phelippeau\inst{2}\orcidlink{0009-0000-7503-3331}}

\authorrunning{Martinez-Sanchez, A. et al.}

\institute{Department of Information and Communications Engineering, University of Murcia, Campus de Espinardo, 30100, Murcia, Spain *\email{anmartinezs@um.es} \and 
Materials and Structural Analysis Division, Advanced Technology, Thermo Fisher Scientific, Bordeaux, France \and
Department of Engineering and Computers Technology, Applied Mathematics, University of Murcia, Campus de Espinardo, 30100, Murcia, Spain}

\maketitle

\begin{abstract}
Object detection is a main task in computer vision. Template matching is the reference method for detecting objects with arbitrary templates. However, template matching computational complexity depends on the rotation accuracy, being a limiting factor for large 3D images (tomograms). Here, we implement a new algorithm called tensorial template matching, based on a mathematical framework that represents all rotations of a template with a tensor field. Contrary to standard template matching, the computational complexity of the presented algorithm is independent of the rotation accuracy. Using both, synthetic and real data from tomography, we demonstrate that tensorial template matching is much faster than template matching and has the potential to improve its accuracy.
  \keywords{Template matching \and Arbitrary object detection \and Cartesian Tensors \and Tomography}
\end{abstract}

\section{Introduction}
\label{sec:intro}

Template matching (TM) is a reference algorithm for arbitrary object detection in computer vision. In particular, TM is used for detecting instances of arbitrary patterns within images from just an input model of the pattern, the template. Contrarily to machine learning approaches, no prior training is required. TM is based on local normalized cross-correlation computation; thus, it presents a good performance for object detection when the instances of the template in the image are subjected only to translation and rotation transformations, no (or not significant) occlusions or any other deformation. Recently, deep learning (DL) approaches have been proposed for TM task improving its performance when previous conditions are not meet on search images, but their outputs are biased by the training set used and none of them works for volumetric images (tomography) \cite{Rocco2017,Geirhos2018,Vock2019,Gao2024}. 

\begin{figure}[!ht]
\centering
\includegraphics[width=0.45\textwidth]{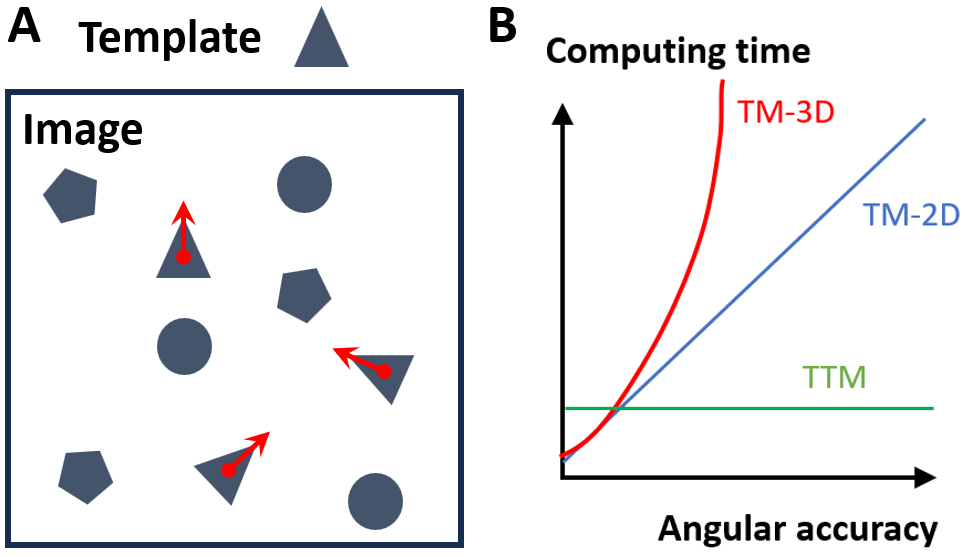}
\caption{TM and its computational complexity. (A) TM consist in detecting positions, center localization, of the instances of an arbitrary input model, the template, in an image, and determining their rotations, or poses. (B) Computing time versus angular accuracy for TM in 2D images (blue) TM in 3D images (red) and the proposed TTM (green).}
\label{fig:tm_vs_ttm}
\end{figure}

As an example, in cryo-electron tomography (cryo-ET) \cite{Fassler2020, Hoffmann2022, Ni2022}, TM is used for macromolecular detection. Cryo-ET is an extension of cryo-electron microscopy (cryo-EM) for three-dimensional (3D) images, or tomograms, which has emerged as the unique technique able to generate 3D representations of the cellular context at sub-nanometer resolution \cite{Turk2020}. TM is used to determine, within an image, the positions and rotations of instances of a structure (typically a macromolecule in cryo-ET) with respect to an input model, or template (see \cref{fig:tm_vs_ttm}.A). Although multiple DL based particle picking algorithms are used for 2D cryo-EM single particle analysis (SPA) \cite{al2019autocryopicker, huang2022weakly, nguyen2021drpnet, bepler2019positive, wagner2019sphire}, its usage in cryo-ET remains very limited. Recently, DL approaches have been proposed to address macromolecular detection task for cryo-ET \cite{Moebel2021, Huang2022, Teresa2023}. However, they present some limitations that make their usage restrained in practice. TM is required for preparing training datasets or processing datasets where the macromolecules are not visually recognizable. Moreover, DL methods cannot retrieve the rotations of the detected instances.

The main limiting factor for TM is its computational cost, especially when working with 3D images and using rotations. TM basic idea is to compute the cross-correlation by an inner product between the template and the image. Correlations are computed in Fourier domain to efficiently handle template translations \cite{Lewis1995, Bohm2000, Roseman2003}. However, this transformation must be repeated for every rotated version of the template. In the case of 2D images, the space of rotations can be identified with the unit circle $\mathbb{S}^1$, 
so its sampling leads to a linear computational complexity $O(N)=O(360/\varepsilon)$, for a given number of samples $N$ determining the angular accuracy $\frac{1}{\varepsilon}$, implying that the maximal distance between the computed rotation and the exact one is smaller than $\varepsilon$ degrees. Furthermore, this complexity increases notably in the 3D case, where the space of rotations is $SO(3)$, so that the complexity raises to $O(N)=O(\left(360/\varepsilon\right)^3)$, which has a cubic dependency on angular accuracy  $\frac{1}{\varepsilon}$. As an example, Reference \cite{Chaillet2023} requires 45123 samples to achieve an angular precision of $\varepsilon=7^{\circ}$.

In this paper, we propose and implement an alternative to classical TM algorithm, called tensorial template matching (TTM), with a computational complexity that is independent of angular accuracy (see \cref{fig:tm_vs_ttm}.B). TTM is based on the recently developed mathematical theory described in \cite{Almira2024}. To do so, TTM integrates into a tensor field, made of symmetric tensors, the information relative to the template over all rotations. This tensor field is computed only once per template. Therefore, TTM allows to find positions and rotations of template instances in an image with a reduced and fixed amount of correlations, just one per each linearly independent tensor component. We experimentally demonstrate that the new mathematical theory behind TTM is correct. Moreover, we also show that TTM is ready to substitute TM for some practical applications by processing real data.

This work uses tensors to encode information over all template rotations, an idea highly related with spherical harmonics. It has been shown that symmetric tensors can be represented using spherical harmonics and vice-versa \cite{Applequist1989}. However, for representing a function of $SO(3)$, the ordinary spherical harmonics are not enough. Hyperspherical harmonics \cite{Dokmanic2009} could be used, but these are fairly complex to handle compared to the simple Cartesian tensors used by the method presented here.

In terms of speeding up TM using cross-correlations, it is possible to reduce the area or volume over which the matching is performed by identifying regions of interest during a preprocessing step \cite{Bohm2000, Dimaio2009}. However, this method risks missing matches if the preprocessing is not entirely accurate and depends on having a suitable preprocessing for a particular template. The method used in \cite{Dimaio2009} finds likely candidates by preprocessing using rotation-invariant features. Afterwards, this method uses a similar basic approach for matching to TTM, considering that tensors and spherical harmonics are equivalent, but with an important difference: TTM does not need to explicitly consider multiple concentric shells, so it allows computing each correlation using the fast Fourier transform.

Steerable filters \cite{Freeman1991} have a similar efficient computational scheme compared to TTM, since for a specific kernel, or template, they use a linear combination of few basis filters to evaluate the convolution at any orientation of the kernel. The design of steerable filters has traditionally been limited to simple kernels \cite{Jacob2004}. Recently, a method has been proposed to generate steerable filters for any arbitrary kernel \cite{Fageot2021}. However, it only offers a practical solution for 2D images. Conversely, our approach solves the problem of TM computational complexity for 3D images. 

\section{Template Matching}
\label{sec:tm}

Here, we introduce TM as it serves as the foundation for the subsequent description of TTM. TM relies on computing the local normalized cross-correlation (LNCC) function of the image, $f$, with respect to the template, $t$, for every voxel $x \in \mathbb{R}^d$ and considering both images elements of $L^2(\mathbb{R}^d)$, where $d$ is the dimension, e.g. in 3D $d=3$. The LNCC can also be seen as a normalized inner product between $f$ and $t$

\begin{equation}\label{eq:lncc}
    c(x,R) = w(x) (f\star t'_R)(x),
\end{equation}
being 
\begin{equation}\label{eq:norm_f}
    w(x)=\frac{1}{\sqrt{\langle S(\tau_x(f))^2, \mathbf{1}\rangle -\frac{\langle S(\tau_x(f)), \mathbf{1}\rangle^2}{M}}}
\end{equation}
the normalization factor for the image. Here, $\tau_x(f)(z)=f(z+x)$ represents the image translation at position $x$ to fit the template center, and $S$ is an operator implemented as
\begin{equation}
    S(f)= m (h\ast f),
\end{equation}
where the mask $m$ equals $1$ within a certain radius around the center and $0$ outside a slightly larger radius; in-between these radii the mask interpolates between $0$ and $1$. $M$ is the sum of voxels in $m$. Operator $S$ also includes the separable filter $h$ whose 1D $z$-transform is $1+a(z+z^{-1}-2)$. For the filter to be a low-pass
filter and positive semidefinite, we should have $0 < a < 1/4$; $a = 1/5$ was chosen for its near isotropic behavior.

The template is zero-normalized by subtracting its mean $\mu=(S(t)\cdot m)/ M$ before the scaling

\begin{equation}\label{eq:norm_t}
    t'=\frac{m(S(t)-\mu)}{\sqrt{\langle S(t)^2, \mathbf{1}\rangle -\frac{\langle S(t), \mathbf{1}\rangle^2}{M}}}.
\end{equation}

The cross-correlation between the image and the template is mathematically expressed as
\begin{equation}\label{correlation}
    (f\star t )(x)=\langle \tau_x(f),t\rangle=\int_{\mathbb{R}^d}f(z+x)t(z)dz.
\end{equation}

In \cref{eq:lncc} the template is rotated by $R$. The whole space of rotations is sampled to generate the final cross-correlation field. The output is a scalar map $c_{R_{opt}}:\mathbb{R}^{d} \rightarrow \mathbb{R}$ with the highest correlation values for each voxel
\begin{equation}
    c_{R_{opt}}(x) = c(x,R_{opt}(x)) ,
\end{equation}
\noindent $R_{opt}(x)$ is the optimal rotation at a given voxel
\begin{equation}
    R_{opt}(x) = \arg \underset{R}{\max}\left \{  c(x,R) \right \} . 
\end{equation}
The highest peaks of $c_{R_{opt}}$ determine the location of different instances of the template in the tomogram. This also means $R_{opt}$ on peaks estimates instances rotations. Accordingly, the density used for sampling the rotations space determines rotation accuracy as well as the detection performance.  

\section{Tensorial Template Matching}
\label{sec:ttm}

In contrast to TM, tensorial template matching, TTM, introduces the concept of tensorial correlation:

\begin{equation}\label{eq:tcc}
    C_n(x) = w(x) (f\star T(t))(x),
\end{equation}

\noindent being $T(t)$ the tensorial template defined as 

\begin{equation}\label{eq:tt}
    T(t)=\int_{SO(d)}R^{\odot n}S(t')_R dR,
\end{equation}

\noindent where $R^{\odot n}$ is a symmetric tensor of degree-$n$ constructed as the $n$ tensor power for rotation $R$. In 3D, $R\in SO(3)$, then rotations can expressed as unit quaternions $q$ with $d'=4$. The \cref{alg:tt} describes how to construct the tensor. This algorithm computes numerically \cref{eq:tt}, which requires to sample uniformly $SO(3)$ space \cite{Alexa2022}.

\begin{algorithm}
\caption{Tensor template generation}\label{alg:tt}
\begin{algorithmic}
    \Require{$t$: template, $m$: mask, $Q$: list of unit quaternions uniformly sampling $SO(3)$}
    \Ensure{$T$: tensorial template}
    \State{Compute $t'$} \Comment{Template normalization}
    \State{$I \gets$ indices independent components degree-$n$ tensor}
    \For{$i\in I$}
        \State{Initialize $T_i$ to a blank array of the size of $t$.}
    \EndFor
    \For{$q\in Q$} \Comment{Integration in $SO(3)$}
        \State{$t_{temp}\gets t'_q$} \Comment{Rotate $t'$ according to $q$}
        \For{$i\in I$}
            \State{$k\gets$ generate component $i$ of $q^{\odot n}$}
            \State{$T_i\gets T_i+kt_{temp}$}
        \EndFor
    \EndFor
\end{algorithmic}
\end{algorithm}

In \cref{eq:tcc}, $C_n:  \mathbb{R}^{d} \rightarrow S^n(\mathbb{R}^{d'})$, $x\hookrightarrow C_n(x)$, is a tensorial field, where $S^n(\mathbb{R}^{d'})$ stands for the space of symmetric tensors of degree-$n$ and dimension $d'$. The key of TTM is that $C_n(x)$ does not depend on rotations $R$, the information for the template over all rotations is integrated in the so-called tensorial template, $T(t)$, defined in \cref{eq:tt}. It is important to remark that the numerical computation of the tensorial template requires to sample the whole space of rotations, $SO(3)$ in 3D, but it is only computed once for each template. Afterward, we can use this tensorial template to compute the tensorial field, $C_n(x)$, for any given image $f$. The \cref{alg:ttm} describes how to generate the tensorial field. Notice that correlations are computed in Fourier domain ($\mathcal{F}$).

\begin{algorithm}
\caption{Tensorial field computation}\label{alg:ttm}
\begin{algorithmic}
    \Require{$f$: image, $T$: tensorial template}
    \Ensure{$C_n$: tensorial field}
    \State{Compute $w$} \Comment{Local normalization of $f$}
    \State{$I \gets$ indices independent components degree-$n$ tensor} 
    \For{$i\in I$}
        \State{$C_i\gets w\mathcal{F}^{-1}\left \{ \mathcal{F}(S(f))\mathcal{F}(T_i)^* \right \}$}
    \EndFor
\end{algorithmic}
\end{algorithm}

The tensorial correlation requires as many correlations as linearly independent components of a degree-$n$ tensor, specifically $\binom{n+d'-1}{n}$. In principle, the higher tensor degree-$n$ the better $T(t)$ integrates $t$ information, however the theoretical speed-up of computing $C_n(x)$ over $c_{R_{opt}}(x)$ is reduced. Here, we work with tensors of degree-$n=4$ for processing 3D images, $d'=4$, therefore $C_n(x)$ computation will require $\binom{7}{4}=35$ correlations against the thousands typically used to compute $c_{R_{opt}}(x)$.

Now the challenges are the determination of the optimal rotation for a given position and the localization of the template instances in the image.

\subsection{Optimal rotation}
\label{ssec:rot_ttm}

The main theorem in \cite{Almira2024} states that, if there is a match between $f$ and $t_{R_{opt}}$  at $x$, then, for $n$ even, $C_n(x)\cdot R^{\odot n}$ gets a global maximum when $R=R_{opt}$. In addition, it has also been proven that finding this global maximum corresponds with finding the dominant eigenvalue-eigenvector pair of $C_n(x)$. Therefore, determining the optimal rotation for a given voxel, $x$, can be accomplished using the shifted symmetric higher-order power method (SS-HOPM) introduced in \cite{Kolda2010ShiftedPM}. Initialization is performed using iterative eigendecomposition, as suggested by \cite{Kofidis2001OnTB}, in combination with simply trying $1000$ uniformly distributed quaternions. The latter is especially useful when templates contain symmetries because the iterative eigendecomposition used to fail. The shift used in the SS-HOPM is determined using an eigendecomposition of the square matrix unfolding of the tensor, inspired by the observation that the algorithm converges whenever this matrix is symmetric positive semidefinite \cite{Kofidis2001OnTB}. This shift appears to be more conservative than the one proposed in \cite{RegaliaKofidisHOPM}, while requiring less effort than the adaptive SS-HOPM proposed by the same authors \cite{Kolda2010ShiftedPM}.

\subsection{Instance positions}
\label{ssec:pos_ttm}

Computing the optimal rotation only at the matching positions is crucial to maximize the computational advantage of TTM over TM. In this work, we propose to convert the tensorial field $$C_n:\mathbb{R}^{d} \rightarrow S^n(\mathbb{R}^{d'}), \quad x\hookrightarrow C_n(x)$$ into a scalar field $$\hat{c}_n: \mathbb{R}^{d} \rightarrow \mathbb{R}, \quad x\hookrightarrow \hat{c}_n(x)$$ with the following properties: 
\begin{enumerate}
    \item The computation of $\hat{c}_n(x)$ is not expensive.
    \item Matching positions correspond to local maxima of $\hat{c}_n$, so they can be efficiently determined by a peak detector.
\end{enumerate}
Therefore, the important step is to define a candidate for $\hat{c}_n(x)$ whose local maxima approximate as close as possible the global maxima over rotations of $C_n(x)\cdot R^{\odot n}$, meanwhile their computation is not expensive. 

On one hand, let us take into account some properties of tensors. As described in \cite{Almira2024}, for any symmetric tensor $T$, it is known that:
\begin{equation*} 
\|T\|_{\sigma}=\max_{\|Q\|=1}\left|\langle T,Q^{\odot n}\rangle \right|= \max_{\|Q\|=1}\left| T\cdot Q^{\odot n} \right|,
\end{equation*} 
where $\|T\|_{\sigma}$ denotes the spectral norm of $T$. Thus,  $$\hat{c}^*_n(x)=\|C_n(x)\|_{\sigma}$$ satisfies the second condition but, unfortunately, it fails with the first one since the computation of the spectral norm of a tensor is NP-hard \cite{DBLP:journals/jacm/HillarL13}. Indeed, in \cite{Hu2022ComplexityAC} is shown that if all but two of the $n$ dimensions are fixed, then the spectral norm of an order $n$ tensor can be computed in polynomial time, while it becomes NP-hard if three or more dimensions are taken as input parameters. In any case, computing $\|C_n(x)\|_{\sigma}$ for all positions $x$ with a given (small) precision $\delta>0$ is too expensive for our purposes. 

On the other hand, in \cite{Almira2024} the Frobenius norm $\|T\|_F$ of a tensor was proposed as an alternative to the spectral norm. Here we show that it can be used as an excellent proxy for finding the spatial locations of instances.  Indeed, we know that $C_n(x)\in S^n(\mathbb{R}^{d'})$ and  (see, e.g., \cite{CaoExtreme}, \cite{Tonelli2022})
\[
\|T\|_{\sigma}\geq \|T\|_F\frac{1}{2^{n-1}}
\]

Thus, if $\|T\|_F$ is large, the spectral norm of $T$ is also large. It follows that a good candidate for the scalar field $\hat{c}_n$ which satisfies the first condition and approximates the second is $$\hat{c}_n(x)=\|C_n(x)\|_{F}.$$ 

For each position identified as a potential peak, the SS-HOPM algorithm is then used to find the exact dominant eigenvalue (and its corresponding eigenvector). 


\begin{figure*}[t]
\centering
\includegraphics[width=1\textwidth]{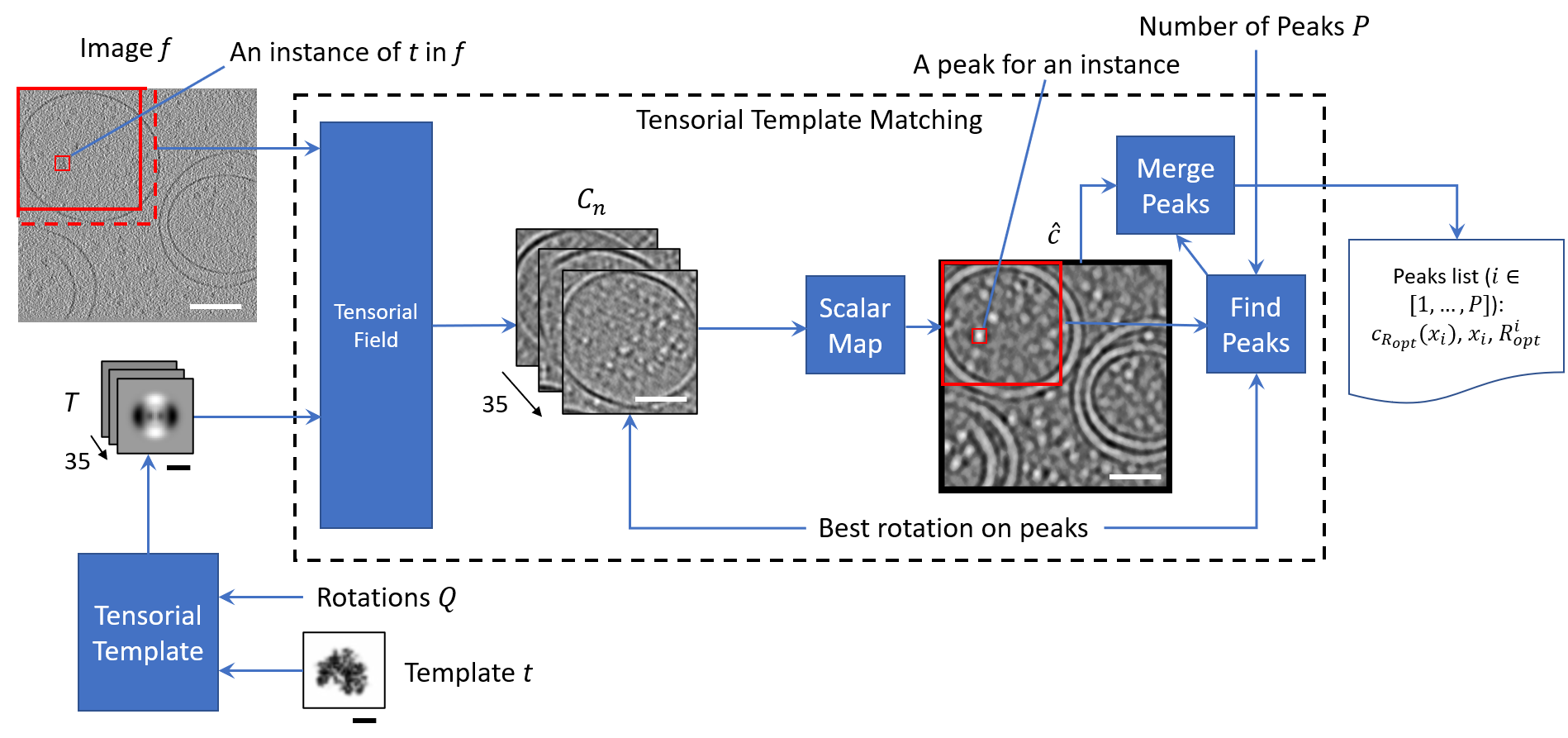}
\caption{TTM implementation scheme. Images represents 2D slices of tomograms, the input tomogram was taken from SHREC dataset \cite{Gubins2021}, and the template was generated from \href{https://www.rcsb.org/structure/5MRC}{PDB-5MRC} atomic model. Black scale bars $200$ {\AA}, white bars $2000$ {\AA}.}
\label{fig:scheme}
\end{figure*}

\subsection{Positions refinement}
\label{ssec:pos_ref_ttm}

As mentioned above and shown experimentally in \cref{sec:res}, the Frobenius norm is a fast proxy for finding positions, but the exact match may be shifted a bit. When positions are shifted, the conditions for finding the optimal rotation are not met (see \cref{ssec:rot_ttm}), so rotation determination is considerably degraded. Consequently, we propose here a heuristic to ensure that this potential shifting is corrected, which is so called TTM-refined (TTM-ref). First, for each peak found, we define a sphere around it with a radius of $r_s$ voxels. Second, we iterate over these voxels and compute the optimal rotation as described in \cref{ssec:rot_ttm}. Finally, the LNCC at just the optimal rotation, $c(x,R_{opt})$, is computed at every voxel returning the position $x$ that attains the highest value. More details in \cref{alg:ttm-r}.

\begin{algorithm}
\caption{Position refinement for TTM (TTM-ref)}\label{alg:ttm-r}
\begin{algorithmic}
    \Require{$f$: image, $t$: template, $C_n$: tensorial field, $L=\{x^i\}_{i=\{1,...,P\}}$: list of $P$ peaks obtained from $\hat{c}_n$, $r_s$: sphere radius in voxels.}
    \Ensure{$L_{ref}=\{(\tilde{x}^i,R^i_{opt})\}_{i=\{1,...,P\}}$: list of $P$ peaks with refined positions}
    \For{$i \in P$} \Comment{Loop for peaks}
        \State{$V_{x^i} \gets$ all voxels in $\{ ||x^i - x|| \leq r_s \} \setminus x^i$}
        \State{$R_{opt} \gets$ optimal rotation from $C_n(x^i)$}
        \State{$c \gets w(x^i)(f\star t'_R)(x^i)$} \Comment{LNCC}
        \For{$j \in V_{x<i}$} 
            \State{$R \gets$ optimal rotation from $C_n(x^j)$}
            \State{$\tilde{c} \gets w(x^j)(f\star t'_R)(x^j)$} 
            \If{$\tilde{c} > c$}
                \State{$c \gets \tilde{c}$}
                \State{$\tilde{x} \gets x^j$}
                \State{$R_{opt} \gets R$}
            \EndIf
        \EndFor
        \State{$L^i_{ref} \gets (\tilde{x},R_{opt})$}
    \EndFor
\end{algorithmic}
\end{algorithm}

TTM-ref is simple but effective, in \cref{sec:synth_res} it is shown that $r_s=3$ voxels is enough for correcting the shifts of the templates presented here. Contrarily to the TM algorithm, where computing the LNCC requires to sample the whole space of rotations and applied to all image positions. Here, the LNCC is only computed once per voxel at the optimal rotation predicted by TTM, and just at the neighborhood of the pre-detected positions. For the sake of efficiency, LNCC is now computed in the real space for every position by extracting a centered local patch of the image with the size of the template.

\subsection{Implementation details}
\label{ssec:imp_details}

Integrating properly the needle over all rotations is a critical part of the procedure. In this work, this is accomplished by summing over a large sampling set of rotations. In the traditional approach, the exact distribution of rotations is not terribly critical, as we only retain the maximum response. However, when using the samples for integration, it is imperative that they are as uniformly distributed as possible, therefore we use the algorithm presented in \cite{Alexa2022}.

TTM has been implemented for processing 3D images, tomograms, following the scheme depicted in \cref{fig:scheme}. The blue boxes represent algorithms. \textit{Tensorial Template} is \cref{alg:tt} and \textit{Tensorial Field} is \cref{alg:ttm}. \textit{Scalar Map} corresponds to compute $\hat{c}_n(x)$. \textit{Find Peaks} is implemented by taking the $P$ highest values of $\hat{c}_n(x)$, but considering a minimum exclusion distance among peaks in accordance with the template diameter, two times the mask radius. The implementation of \textit{Find Peaks} also includes the heuristic for position refinement described in \cref{ssec:pos_ref_ttm}. A block processing scheme is proposed to avoid having many full copies with the dimensions of the original tomogram loaded in the main memory. For example, in cryo-ET the typical dimensions of a tomogram are around 1000x1000x250 voxels. Overlapping halos with template radius length is needed to prevent unreliable correlations on block borders. \textit{Merge Peaks} gathers all peaks found in all blocks and sorts them to filter the $P$ highest peaks in the whole image. The final output is a list of $P$ template instances ranked by their similarity in terms of the LNCC with respect to the input template and including their pose, i.e., their rotation with respect to the input template. Finally, the user determines which peaks are real peaks just by defining a minimum threshold for the similarity.

\section{Experimental results}
\label{sec:res}

All experiments are performed on an Ubuntu 20.04 workstation with a CPU Intel Xeon W2255 10-cores and 32GB RAM. In addition, TM implementation was accelerated with an Nvidia RTX4090 GPU.

\begin{figure*}
\centering
\includegraphics[width=1\textwidth]{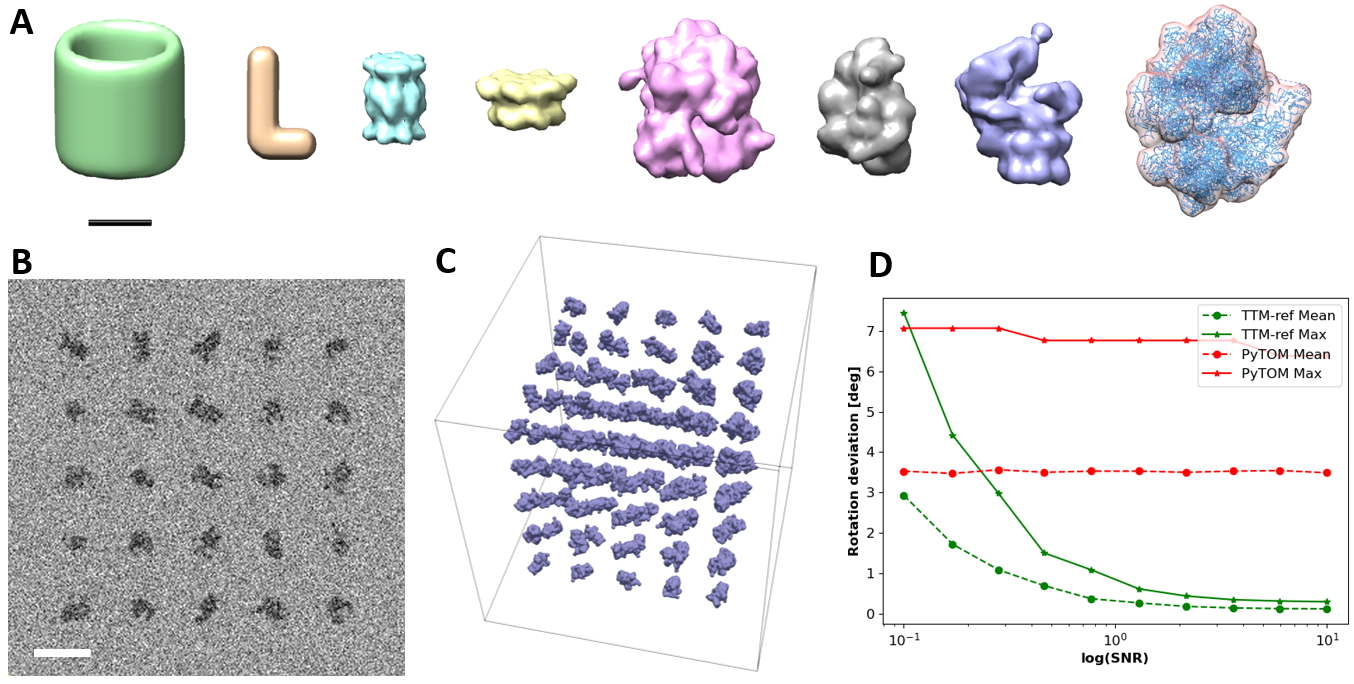}
\caption{Synthetic data. (A) Isosurfaces of the 3D templates with 10 {\AA} voxel size, from left to right: cylinder, L-shape, 3J9I, 3CF3, 4V4R, 1QvR, 4CR2 and 5MRC. The cylinder shows radial symmetry along the axis, 3J9I D7 and 3CF3 C6. 5MRC isosurface has transparency and also shows the atomic model used to generated the template density. Scale bar 100 {\AA}. (B) 2D slice on XY-plane of the tomogram with SNR=0.1 containing instances of 4CR2 at random rotations. Scale bar 500 {\AA}. (C) Isosurface of a noise free tomogram with 4CR2 instances. (D) Deviation in the rotations for the 4CR2 template at different $SNR$ levels. Horizontal axis is in log scale. }
\label{fig:synth_data}
\end{figure*}

\subsection{TTM determines accurately positions and rotations}
\label{sec:synth_res}

To validate the algorithms and quantitatively analyze their accuracy, we have prepared some synthetic tomograms. We have created 8 tomograms, one per each template used in this study. Two templates are simple shapes, a cylinder and a L-shape, and the other six are taken from atomic models of the \href{https://www.rcsb.org/}{Protein Data Bank} generating 3D densities with 10 {\AA} voxel size (see \cref{fig:synth_data}.A). The proteins have been selected to represent different sizes and shapes, within the Protein Data Bank each entry is identified by a four digits code. Additionally, some templates present different symmetries. Each tomogram contains a 5x5x5 grid with 125 randomly rotated copies of the same template. The shape of all templates is cubic and has an odd size in order to generate templates with unambiguous centers. The size is derived from the bounding box of the structure, such that the center of the template corresponds to the center of mass of the structure, and the size of the template comprises the diagonal of the structure’s bounding box ensuring that any rotation fits into it. An example of a tomogram is showed in \cref{fig:synth_data}.B-C.

The tomograms were processed with the two implementations proposed here, TTM and TTM-ref. For comparison, the tomograms were also processed with 
the widely used software in cryo-ET for TM, PyTOM (v1.1, Aug 29, 2023) \cite{Hrabe2012, Gubins2021, Chaillet2023}. 

Position accuracy is computed as the Euclidean distance in voxels between actual particle positions and algorithms output (see \cref{tab:pos}). The distance between two rotations expressed as quaternions, $q_1$ and $q_2$, is measured by the angle $2\arccos \left( |q_1 \cdot q_2| \right )$ in degrees \cite{Huynh2009} (see \cref{tab:rot}).

The results in \cref{tab:pos} show that, in general, the Frobenious norm is a good proxy for position detection since for most of the templates it achieves subvoxel accuracy, and in the rest, its shifting is always under 3 voxels. Regarding rotations, when positions are perfectly detected, TTM obtains an accuracy below $0.3^{\circ}$. Conversely, rotation accuracy of TM is limited by the sampling of $SO(3)$, in the case of 
PyTOM using 45123 samples, the results for maximum distances ranged $4^\circ$ to $13^\circ$. 

We observed that errors detecting positions greatly spoils rotation determination by TTM as advanced in \cref{ssec:pos_ttm}. However, TTM-ref, TTM with position refinement (see \cref{ssec:pos_ref_ttm}), is able to compensate errors in positions, enabling also a highly accurate rotation determination for all templates. According with the results obtained, TTM accuracy is not affected by symmetry.  


\begin{table*}
\centering
 \caption{Positions precision. Values are mean / maximum Euclidean distance in voxels between actual and predicted positions.}
\begin{tabular}{@{}l|c|c|c|c@{}}
\toprule
                & \textbf{Cylinder} & \textbf{L-shape}  & \textbf{3J9I} & \textbf{3CF3}  \\ \midrule
 \textbf{PyTOM (TM)}    &    0.32 / 1.73   &   2.16 / 3.74   &  2.35 / 3.46  &  2.23 / 3.46   \\
 \textbf{TTM}   &   2.38 / 2.83  & 0.0 / 0.0  &  0.0 / 0.0 &  0.0 / 0.0   \\
 \textbf{TTM-ref}  &   0.0 / 0.0    & 0.0 / 0.0  &  0.0 / 0.0 &  0.0 / 0.0   \\ \bottomrule

         & \textbf{4V4R} & \textbf{1QVR} & \textbf{4CR2} & \textbf{5MRC}   \\ \midrule
     \textbf{PyTOM (TM)}  &  2.19 / 3.74  &  2.13 / 3.74 &  2.15 / 3.74  &  2.19 / 3.74  \\
     \textbf{TTM}  &    0.0 / 0.0   &  0.0 / 0.0 &  0.21 / 1.0 &  0.0 / 0.0   \\
     \textbf{TTM-ref} &    0.0 / 0.0  &  0.0 / 0.0 &  0.0 / 0.0  &  0.0 / 0.0  \\ \bottomrule
 
\end{tabular}
 \label{tab:pos}
\end{table*}

\begin{table*}
\centering
 \caption{Rotations precision.  Values are mean / maximum angular distance in degrees between actual and predicted rotations.}
\begin{tabular}{@{}l|c|c|c|c@{}}
\toprule
                & \textbf{Cylinder} & \textbf{L-shape}  & \textbf{3J9I} & \textbf{3CF3}  \\ \midrule
 \textbf{PyTOM (TM)}    &    2.20 / 4.09 &   5.70 / 12.56  & 2.48 / 4.69 &  2.51 / 5.14    \\
 \textbf{TTM}   &    0.03 / 0.06 &   31.40 / 39.01 & 0.03 / 0.12 &  0.06 / 0.17   \\
 \textbf{TTM-ref} &    0.03 / 0.06 &   0.03 / 0.24   & 0.03 / 0.12 &  0.06 / 0.17  \\ \bottomrule

  \textbf{PyTOM (TM)}  & \textbf{4V4R} & \textbf{1QVR} & \textbf{4CR2} & \textbf{5MRC}   \\ \midrule
  \textbf{PyTOM (TM)}    &  3.22 / 5.27 &  3.84 / 7.07 &  3.49 / 6.39 &  3.02 / 5.20   \\
  \textbf{TTM}  &  0.14 / 0.26 &  0.09 / 0.24 &  0.97 / 7.60 &  0.07 / 0.16 \\
 \textbf{TTM-ref} &  0.14 / 0.26 &  0.09 / 0.24 &  0.11 / 0.26 &  0.07 / 0.16   \\ \bottomrule
 
\end{tabular}
 \label{tab:rot}
\end{table*}

In \cref{fig:synth_data}.D, we add additive Gaussian noise to a synthetic tomogram (protein 4CR2) to analyze its impact on TTM and PyTOM performance. Deviations in rotations are under $1^\circ$ for $SNR>1$ and reach a maximum around $7^\circ$ in the worst case for $SNR=0.1$, mean deviations are under $3^\circ$, and under PyTOM values, up to $SNR=0.1$. However, TTM seems to be more sensitive to noise than PyTOM. Positions are not displayed because they are always perfectly detected after refinement. 

\subsection{TTM outperforms TM for ribosome localization in cryo-ET}
\label{sec:exp_res_pos}

To evaluate the accuracy of TTM for object detection on real 3D images, specifically for ribosome detection on a set of cryo-tomograms, we have taken the dataset publicly available in \href{https://www.ebi.ac.uk/empiar/EMPIAR-10988}{EMPIAR-10988} \cite{Teresa2023}. This dataset contains 10 tomograms of \textit{Schizosaccharomyces pombe} specimens at 13.68 {\AA}. We took this dataset because it also includes a list with ribosome positions laboriously generated by experts, contrarily to other public datasets in cryo-ET. Currently, there is no method available to precisely identify the entire population of a macromolecule. TM is the only alternative to provide an estimation. However, these data have undergone careful processing and examination by experts in the field. The list of ribosome positions provided in this dataset was obtained through an initial particle picking using TM (PyTOM) with an intentional over-picking followed by manual refinement. In a meticulous and labor-intensive process, experts thoroughly review each individual instance to eliminate false positives and carefully examine the tomograms to avoid missing instances. Moreover, the authors were able to obtain several high-resolution averages from the instances finally isolated. Consequently, we can be confident that positions provided in this dataset actually correspond to real ribosome instances and represent quite well the whole population, so we can use them as ground truth.

\begin{figure*}[!ht]
	\centering
	\includegraphics[width=1\textwidth]{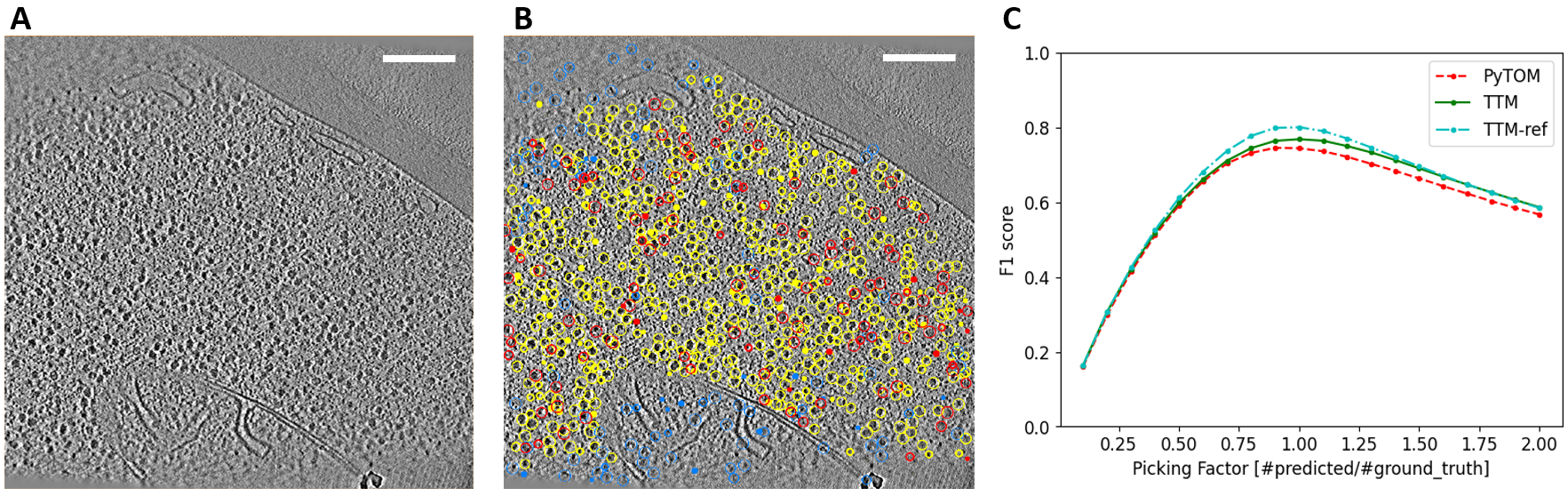}
	\caption{Object detection in a real dataset. (A) A 2D slice on XY-plane of a real tomogram containing a region of a \textit{S. pombe} cell. (B) Overlaid on the 2D slice the true positives (yellow), false positives (red) and false negative (blue) detected by TTM. (C) TTM and PyTOM global F1-scores for the dataset \href{https://www.ebi.ac.uk/empiar/EMPIAR-10988/}{EMPIAR-10988}. (A-B) Scale bars 2000 {\AA}.}
	\label{fig:exp_data}
\end{figure*}

\begin{figure}[!ht]
	\centering
	\includegraphics[width=1\textwidth]{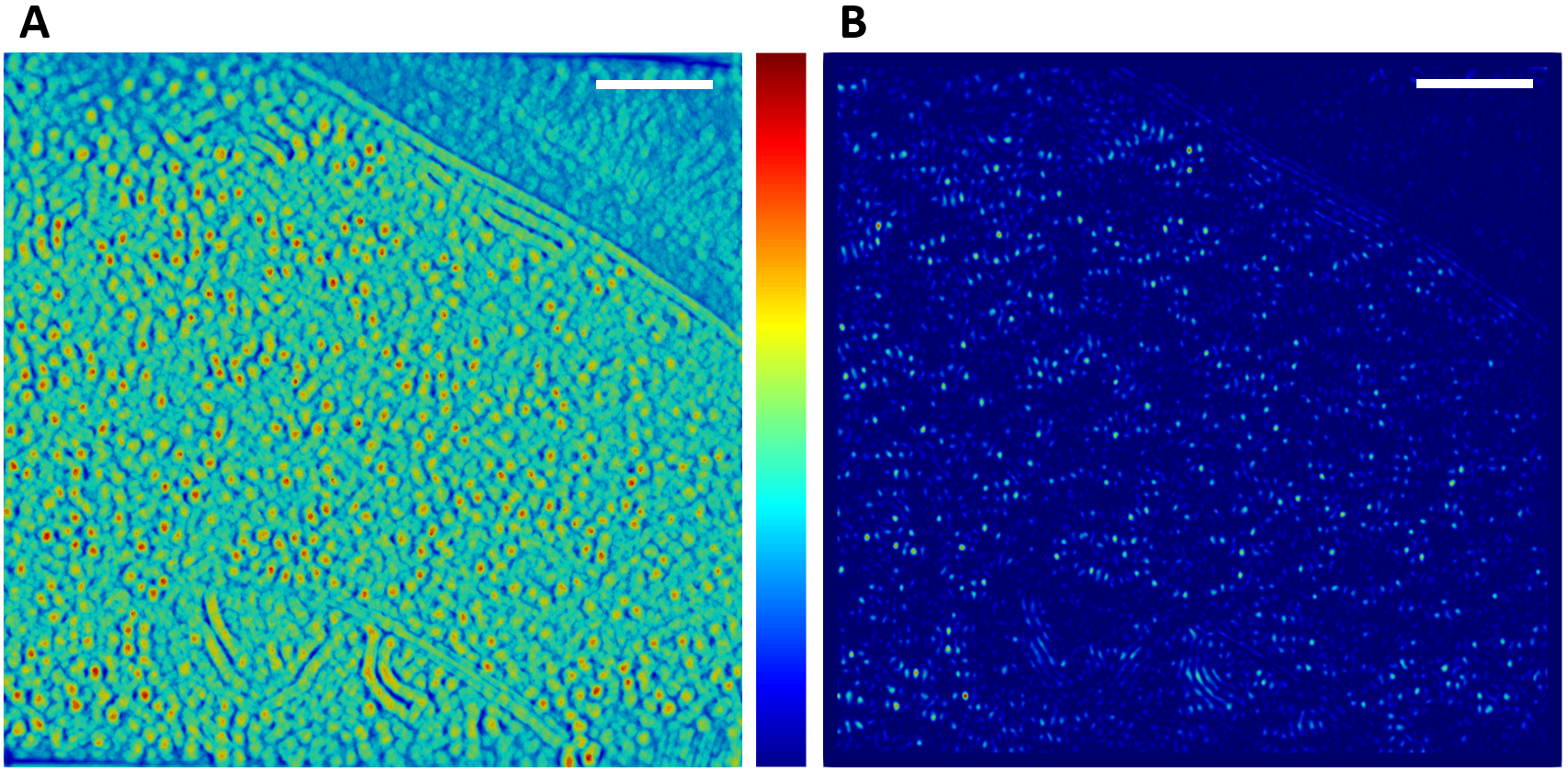}
	\caption{Correlations maps. (A) 2D slice of the tomogram in \cref{fig:exp_data}.A with TM cross-correlation $c_{R_{opt}}$. (B) TTM Frobenious norm $\hat{c}_n$. (Center) Colormap. Scale bars 2000 {\AA}. }
	\label{fig:cmaps}
\end{figure}

\begin{figure}[!ht]
	\centering
	\includegraphics[width=1\textwidth]{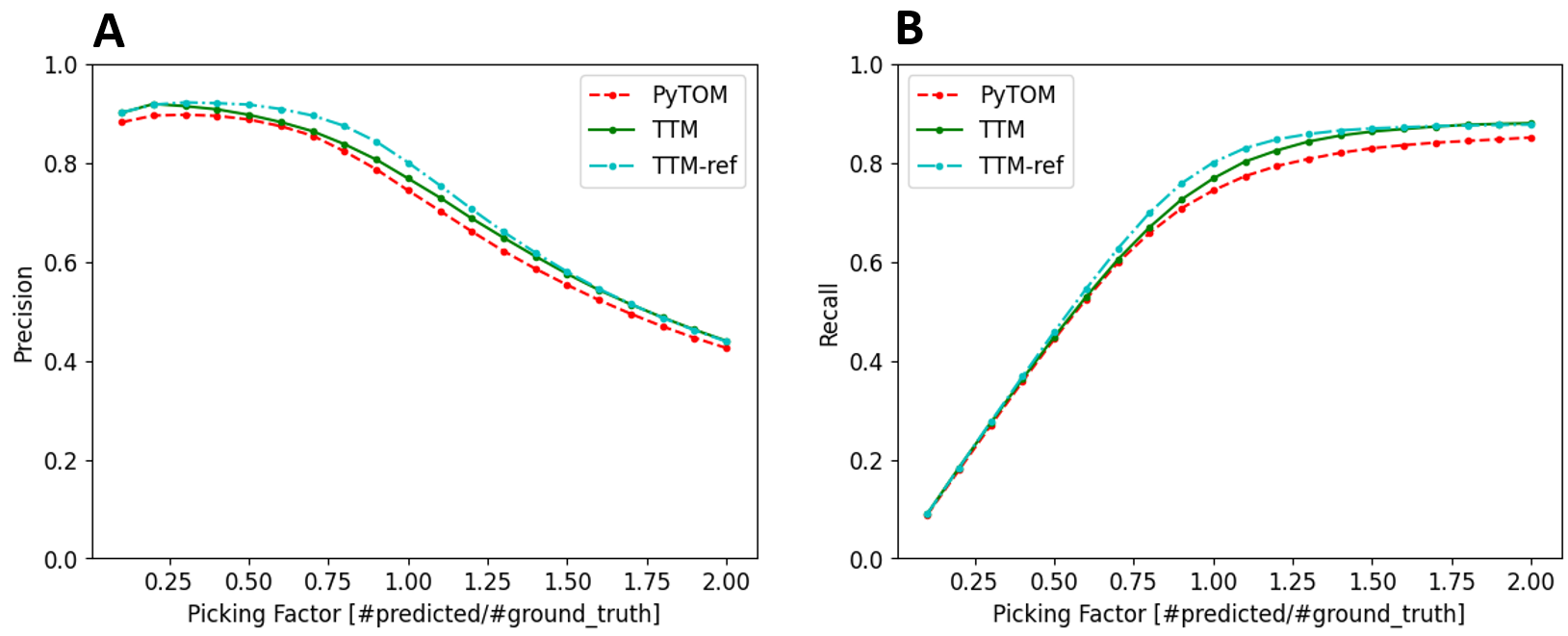}
	\caption{Object detection performance. (A) Precision for processing the dataset \href{https://www.ebi.ac.uk/empiar/EMPIAR-10988/}{EMPIAR-10988}. (B) Recall. }
	\label{fig:pre_rec}
\end{figure}

\cref{fig:exp_data} shows TTM performance for ribosome detection in \href{https://www.ebi.ac.uk/empiar/EMPIAR-10988/}{EMPIAR-10988} dataset with respect to the ground truth and in comparison with TM. We use PyTOM as TM matching implementation, because it is nowadays the most popular in cryo-ET and has been largely improved during years, since its initial presentation \cite{Hrabe2012} to its most recent version optimized for GPU hardware \cite{Chaillet2023}. Here, we have used the ribosome density \href{https://www.ebi.ac.uk/emdb/EMD-14411}{EMD-14411}, obtained from EMPIAR-10988 dataset by subtomogram averaging, as template. An example of the cross-correlation map provided by TM, $c_{R_{opt}}$, and the Frobenious norm given by TTM, $\hat{c}_n$, is available in Supp. Mat. (see \cref{fig:cmaps}). To evaluate the detection performance, we compute the F1-score (harmonic mean of the precision and recall), assuming that a correct detection is a predicted peak closer than 10 voxels (approximately the ribosome radius) to a position in the ground truth list. TTM F1-score approximates 0.8 when the number of peaks nearly corresponds with the number of ground truth instances. That is, when picking factor, ratio between the number of predicted peaks and the ground truth, is 1. TTM-ref is a bit better but not so different. Notice that TTM exceeds PyTOM for the whole curve. Graphs for precision and recall are available in the Supp. Mat. (see \cref{fig:pre_rec}). 

Recently, DL methods have been developed for macromolecular detection in cryo-ET, but they are limited to scenarios where reliable annotations are available for training and, unlike TM and TTM, none is able to determine the rotations for every detected instance. Conversely, TTM has been designed to be an alternative to TM allowing to process datasets when only an input template is available. Therefore, it is not suitable to compare TTM with DL methods trained from exactly the same real dataset used for prediction. Instead, we take DeepFinder \cite{Moebel2021}, the best performer in terms of macromolecular localization along the different editions of SHREC (Classification in Cryo-electron Tomograms) track \cite{Gubins2019,Gubins2021}. Then, we train it with 10 fabricated tomograms including randomized (translations and rotations) copies of the ribosome template within realistic cellular contexts \cite{Martinez-Sanchez2023}. In addition, we also adapted DeepFinder to provide an output like TM and TTM cross-correlation enabling peaks extraction for different thresholds. Results are pretty similar to those obtained by TM and TTM (see Supp. Mat. \cref{tab:dl}) but DeepFinder (or any other DL approach) cannot assign rotations.


\begin{figure}[!ht]
	\centering
	\includegraphics[width=0.5\textwidth]{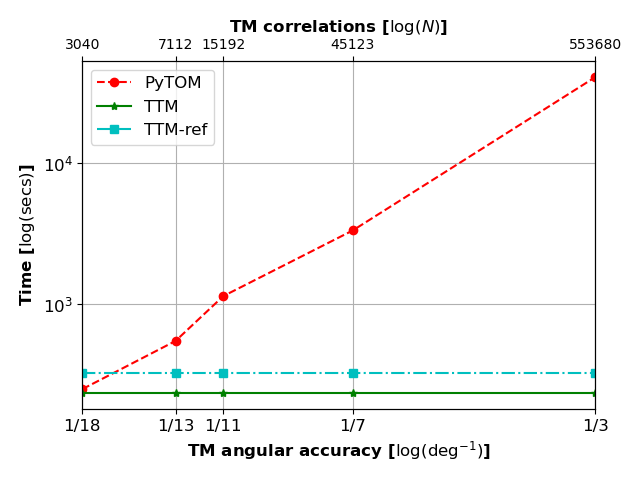}
	\caption{Running times. Comparison between our implementation of TTM and a well-known GPU implementation of TM (PyTOM) for processing a single \href{https://www.ebi.ac.uk/empiar/EMPIAR-10988/}{EMPIAR-10988} tomogram. Axes are in log scale.}
	\label{fig:times}
\end{figure}

\subsection{TTM is faster than TM GPU implementations}
\label{sec:exp_res_time}


\cref{fig:times} is the experimental counterpart of the complexity analysis presented in the introduction section of this paper (see \cref{fig:tm_vs_ttm}). \cref{fig:times} shows the running times in log scales for processing a single tomogram of the real dataset used in \cref{sec:exp_res_pos}. The effective dimensions, excluding padding voxels added during 3D reconstruction, of this tomogram are 960x928x230 voxels. Here, we compare the running times of the TTM implementations (with and without position refinement) against PyTOM, which provides different samplings of the rotations space obtaining different angular precision ranging from $3^{\circ}$ to $18^{\circ}$. Despite PyTOM has recently been optimized to run on GPU architectures \cite{Chaillet2023}, it still requires processing times of several hours to achieve an angular precision below $10^{\circ}$ because of the cubic computational complexity dependency with angular accuracy, $O((360/\epsilon)^{3})$. Conversely, and without GPU acceleration, TTM takes less than four minutes to process a tomogram, 90 seconds more if the refinement is activated. TTM running time is independent of the angular accuracy, as its computational complexity is $O(1)$. TTM only uses the CPU, where the 35 correlations are distributed across various CPU cores as well as the computation of the eigenvectors on the detected peaks. 

Generating the tensorial template is computationally costly (see \cref{alg:tt}), it took 16 minutes for the ribosome template. But this is not relevant for processing time because tensorial template generation is independent of the tomograms to process.



\section{Discussion}

In this paper, we propose and implement a new computer method, which comprises several algorithms for the fast computation of the normalized correlation considering at the same time translation and rotation of the template. The method is based on the main theorem proposed and demonstrated in \cite{Almira2024} but adapted and extended here for processing (tomograms). Our experimental results validate the novel mathematical approach based on Cartesian tensors.

We experimentally demonstrate that TTM is more accurate than TM for predicting positions and rotations in tomograms free of artifacts. Moreover, because of the complexity reduction of TTM with respect to TM, processing times for TTM are significantly lower than TM without the necessity of relying on GPU hardware for acceleration. Nowadays, the application of TM for tomography has basically been limited to cryo-ET, so we studied the performance of TTM in this domain. Now, TTM solves efficiently the problem of arbitrary object detection, besides rotation assignment, from just an input template, therefore TTM can also be used in application domains where TM computational cost was the limiting factor. 

Furthermore, we demonstrate that a degree-$4$ tensor was sufficient to recover rotation information accurately in practice. As a future work, we plan to study in detail the impact of the tensor degree in the robustness against image distortions.

\section*{Acknowledgments}

A.M-S. was supported by the Ramon y Cajal Program of the Spanish State Research Agency (AEI) through the MICIU/AEI/10.13039/501100011033 and the European Union NextGenerationEU/PRTR under Grant RYC2021-032626-I and by the University of Murcia through the Attract-RYC 2023 Program.

%
%
\bibliographystyle{splncs04}
\bibliography{main}


\newpage

\section{Additional figures and tables}
\label{sec:supp}

\begin{table*}[!ht]
	\centering
	\caption{Detection performance in a real dataset. Left results for the best performing tomogram in \href{https://www.ebi.ac.uk/empiar/EMPIAR-10988/}{EMPIAR-10988}, right for the worst.}
	\begin{tabular}{@{}l|c|c|c@{}}
		\toprule
		& \textbf{F1-Score} & \textbf{Precision}  & \textbf{Recall} \\ \midrule
		\textbf{PyTOM (TM)}   & 0.82 / 0.67 & 0.87 / 0.81 & 0.78 / 0.57 \\
		\textbf{TTM}          & 0.80 / 0.64 & 0.80 / 0.72 & 0.80 / 0.58 \\
		\textbf{TTM-ref}      & 0.85 / 0.67 & 0.85 / 0.75 & 0.85 / 0.60 \\
		\textbf{DeepFinder}  &   0.84 / 0.61    & 0.89 / 0.69  &  0.8 / 0.56    \\ 
		\bottomrule
		
	\end{tabular}
	\label{tab:dl}
\end{table*}

\end{document}